\documentclass[10pt,twocolumn,letterpaper]{article}

\usepackage{cvpr}              

\definecolor{cvprblue}{rgb}{0.21,0.49,0.74}
\usepackage[pagebackref,breaklinks,colorlinks,allcolors=cvprblue]{hyperref}

\usepackage{multirow}
\usepackage[accsupp]{axessibility}  

\def\paperID{8954} 
\def\confName{CVPR}
\def\confYear{2025}

\title{SparseAlign: A Fully Sparse Framework for Cooperative Object Detection}

\author{
    Yunshuang Yuan$^{1}$ \quad
    Yan Xia$^{2,3 \footnotemark[2]}$ \quad
    Daniel Cremers$^{2,3}$ \quad
    Monika Sester$^{1}$ \\
    {\small$^{1}$Leibniz University Hannover \quad
    $^{2}$Technical University of Munich \quad
    $^{3}$Munich Center for Machine Learning (MCML)
    }\\
    {\tt\small \{yunshuang.yuan, monika.sester\}@ikg.uni-hannover.de} \quad
    {\tt\small \{yan.xia, cremers\}@tum.de}
}

\begin{document}
\maketitle

\let\oldthefootnote\thefootnote
\renewcommand{\thefootnote}{\fnsymbol{footnote}} 
\footnotetext[2]{Corresponding author.} 
\let\thefootnote\oldthefootnote
\begin{abstract}
Cooperative perception can increase the view field and decrease the occlusion of an ego vehicle, hence improving the perception performance and safety of autonomous driving. Despite the success of previous works on cooperative object detection, they mostly operate on dense Bird's Eye View (BEV) feature maps, which are computationally demanding and can hardly be extended to long-range detection problems. More efficient fully sparse frameworks are rarely explored. In this work, we design a fully sparse framework, \textit{SparseAlign}, with three key features: an enhanced sparse 3D backbone, a query-based temporal context learning module, and a robust detection head specially tailored for sparse features. Extensive experimental results on both OPV2V and DairV2X datasets show that our framework, despite its sparsity, outperforms the state of the art with less communication bandwidth requirements. In addition, experiments on the OPV2Vt and DairV2Xt datasets for time-aligned cooperative object detection also show a significant performance gain compared to the baseline works.
\end{abstract}

\section{Introduction}
\label{sec:intro}

Cooperative Perception (CP) for autonomous driving has gained significant attention due to its potential to enhance road safety. 
Sharing Collective Perception Messages (CPMs) among Intelligent (traffic) Agents (IAs) in the vehicular network effectively reduces occlusions and expands the field of view, improving perception performance. In many recent works~\cite{fcooper,Xu2022opv2v,fpvrcnn,Wang2020v2vnet,Yu2022dairv2x,v2vam,Yuan_gevbev2023}, CP has proven to be effective for different tasks, such as cooperative object detection~\cite{fcooper,Wang2020v2vnet,fpvrcnn,Xu2022opv2v}, Bird's Eye View (BEV) semantic segmentation~\cite{xu2022cobevt,Yuan_gevbev2023}, and 3D occupancy semantic segmentation~\cite{song2024collaborative}. In this work, we focus on LiDAR-based cooperative vehicle detection.

Training a cooperative 3D object detector is computationally demanding as it needs to process data from multiple vehicles, especially when the model is built to digest sequential data from several subsequent frames to learn the temporal context. Following Fcooper~\cite{fcooper}, recent works~\cite{Xu2022opv2v,Yu2022dairv2x,Xu2023_v2vreal,Yin2023v2vformer_plus,Cui2022coopernaut,He2023v2x_ahd,Xu2022v2xvit} encode the 3D point cloud into dense BEV feature maps for further processing, sharing, and fusion. However, the computational cost of this pipeline increases quadratically with the perception range. Although cooperation among IAs can improve long-range detection performance, the IAs are normally not in the best positions to optimize the performance gain. The long-range detection performance of the local detector directly influences the overall performance. Besides, sharing the BEV feature maps among IAs requires extensive communication resources.
Instead, the sparse operations leverage the sparsity of point clouds and have only linear computational complexity relative to the number of points. In addition, the output sparse features can be more efficient for sharing.
It is therefore beneficial to perform cooperative object detection based on sparse operations only.

\begin{figure}[t]
    \centering
    \includegraphics[width=\linewidth]{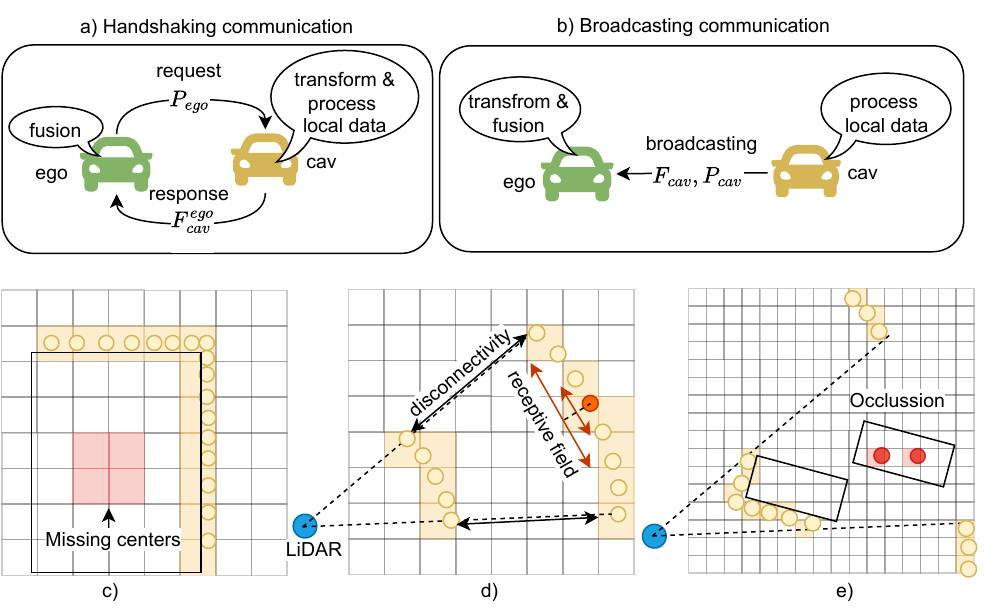}
    \caption{a,b) Communication strategies for cooperative object detection. c-e) Issues of sparse convolutional backbone networks. c) Center Feature Missing (CFM). d,e) Isolated Convolution Field (ICF) caused by ring disconnectivity and occlusion.}
    \label{fig:sparse_issues}
\end{figure}

However, building a fully sparse framework to compete with the dense ones is non-trivial. The first unavoidable issue is \textbf{Center Feature Missing}~\cite{fan2022fully} (CFM, shown in \cref{fig:sparse_issues} c)). Due to the range view of on-board LiDARs, there are usually no scan points in the vehicles' center area. However, compared to the edge points, the center points have a better ability to represent the whole object. Lacking center points for learning, the sparse frameworks perform worse than the dense ones that directly learn the features for center points. The second issue is the poor connectivity between points scanned by different laser beams. As shown in \cref{fig:sparse_issues} d), this disconnectivity issue in distant areas may lead to isolated voxel blobs; the receptive field only enlarges to the blob scale as the convolutional layers go deeper. The isolated voxel blobs never exchange information with each other, limiting the backbone's capability of capturing global features. We call this issue \textbf{Isolated Convolution Field} (ICF).
It also happens in the occlusion areas as shown in \cref{fig:sparse_issues} e): although the occluded vehicle has very few scan points (red), it is expected to be detected based on the neighboring vehicles and the clues in the previous frames. However, ICF introduces difficulties for these isolated points to aggregate information from neighbors and the global environment. 

To overcome the above problems, we build a \textbf{S}parse \textbf{UNet} backbone network \textit{SUNet} with the \textbf{Coordinate-Expandable sparse Convolution} (CEC) implemented in the MinkowskiEngine~\cite{minkunet} library. 
We apply CEC on $4\times$ and $8\times$ downsampled 3D voxels to increase the connectivity between ICFs. On the 2D BEV features, the CECs are used to expand the sparse coordinates, assuring the center features for all scanned vehicles. On top of SUNet, we build \textit{TempAlign Module} (TAM) to learn the temporal context from the historical frames in a query-based~\cite{Wang2023streampetr,streamlts} manner, \textit{PoseAlign Module} (PAM) to correct the relative pose errors between the cooperative and the ego agents, and \textit{SpatialAlign Module} (SAM) to fuse the shared object queries based on the corrected poses.
In addition, we propose the \textit{CompassRose} encoding for the regression of BBox directions to further improve performance on both Cooperative Object Detection (COOD) and the Time-Aligned COOD~\cite{streamlts} (TA-COOD) tasks. 

To summarize, the main contributions of this work are:
\begin{itemize}
    \item We investigate the issues of building a fully sparse network for LiDAR-based COOD and analyze how these issues impact the detection performance.
    \item We propose a novel fully sparse network, \textit{SparseAlign}, via addressing the ICF and CFM problems of the sparse 3D backbone SUNet with CECs. Additionally, we introduce the SAM module, specifically tailored to ensure accurate fusion of sparse features.
    \item We assess the communication bandwidth and precision trade-offs of the different approaches, finding that our proposal reduces bandwidth consumption by up to 98\%, compared to previous methods using dense BEV feature maps for LiDAR-based cooperative object detection.
    \item We conduct extensive experiments on OPV2V, DairV2X, OPV2Vt, and DairV2Xt benchmarks and show that the proposed \textit{SparseAlign} has a significant improvement over state-of-the-art methods. 
\end{itemize}

\section{Related Work}
\label{sec:lite}

\subsection{Cooperative Object Detection (COOD)}
COOD frameworks are classified into sharing CPMs with raw data, intermediate features, and detections. Recent works~\cite{cooper,Yuan2021comap,Wang2020v2vnet,Xu2022opv2v} have shown that intermediate feature sharing has the best performance due to its adjustable and learnable CPMs. We also use this strategy in our framework. In the perspective of communication, there are two strategies: \textit{handshaking} and \textit{broadcasting} as shown in \cref{fig:sparse_issues} a) and b). With handshaking communication~\cite{Xu2022opv2v,He2023v2x_ahd,Xu2022v2xvit,Xu2023_v2vreal,Yuan_gevbev2023,Yin2023v2vformer_plus,streamlts}, the ego vehicle sends a request message that contains its pose to the cooperative IAs so that they can transform the local raw data to the ego coordinate system and then process their data to obtain the features for sharing. This simplifies the feature fusion process, as the features are directly learned in the goal coordinate system. In contrast, the \textit{broadcasting} method~\cite{Wang2020v2vnet,coalign,Yu2023v2xseq,Cui2022coopernaut} directly processes the data in the local coordinate system and transmits it to other IAs. In the fusion stage, the ego vehicle must transform the cooperative features into the ego coordinate system and then conduct the fusion process. 
In comparison, \textit{broadcasting} is more efficient as it does not require a prior communication agreement with the cooperative agents in highly dynamic driving scenarios where the communication partners change frequently. 

\begin{figure*}[t]
    \centering
    \includegraphics[width=\textwidth]{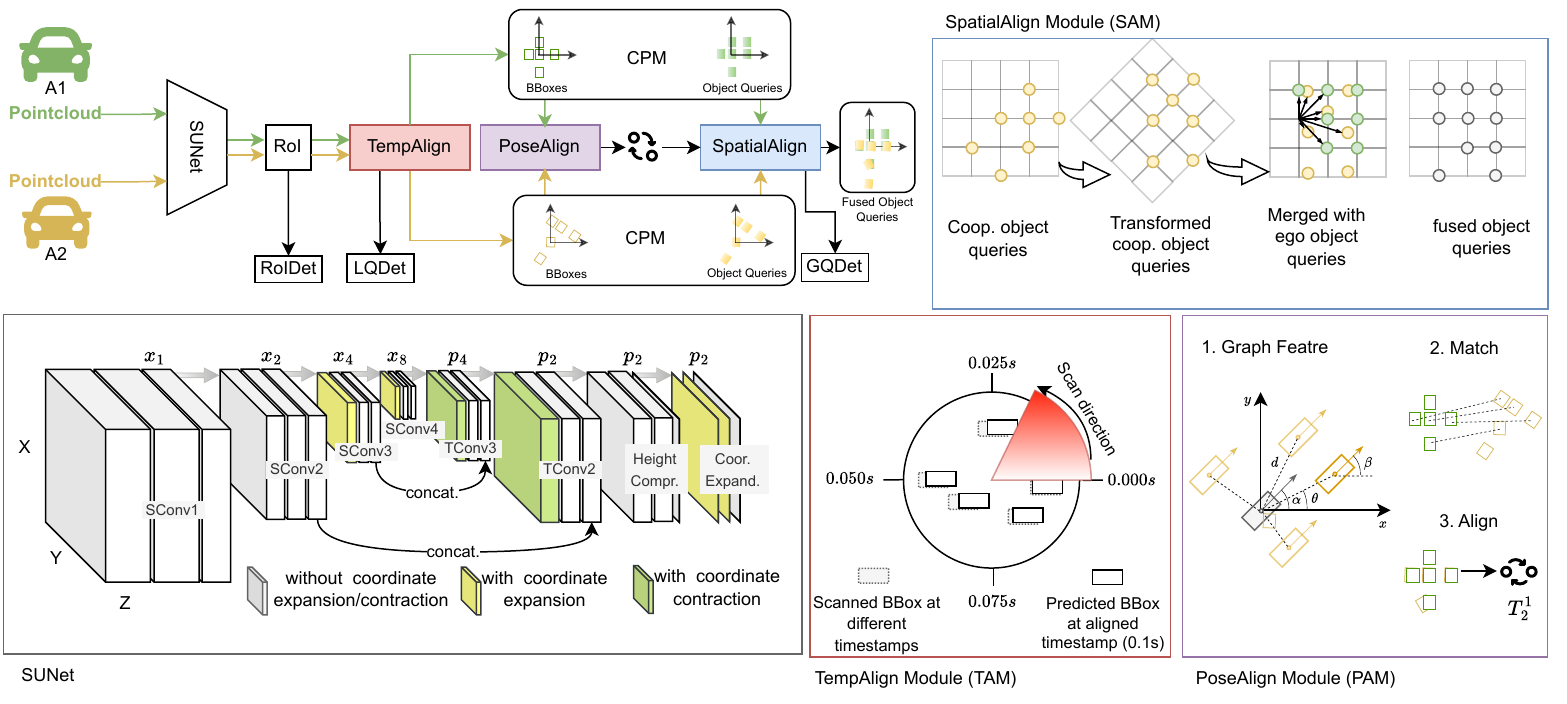}
    \caption{Overview of SparseAlign framework}
    \label{fig:pipeline}
\end{figure*}

Achieving COOD presents several challenges, including communication latency and bandwidth limitations, inaccurate IA poses, and sensor asynchrony.
To reduce the \textbf{communication bandwidth} consumption,
\cite{Xu2022opv2v,Wang2020v2vnet,fcooper,Yin2023v2vformer_plus} compress the learned BEV feature maps to fewer channels via convolution layers, while \cite{Where2comm} utilizes the learned confidence map and \cite{Yuan_gevbev2023} leverages the uncertainty map to select the most important information for sharing. Instead, \cite{fpvrcnn} employs the two-stage detector PVRCNN~\cite{Shi2019PVRCNN} and shares the 3D keypoint features from the second stage. To be more compact, \cite{streamlts,TransIFF,fan2023quest} share the object query features~\cite{Wang2023streampetr} which are highly flexible to blend all relevant information, such as scanning time and IA pose, about an object into one unique feature vector. In this work, we also leverage the object query features to learn the temporal and spatial context of the object. 
\textbf{Pose errors} can dramatically deteriorate the detection results~\cite{Wang2020v2vnet}. To this end, \cite{fpvrcnn,Yuan2022loccor} leverage the predicted Bounding Boxes (BBoxes) and select semantically classified keypoints to align the relative pose between the ego and the cooperative vehicles. Instead, \cite{coalign} solely uses the BBoxes for pose alignment via the \textit{Pose Graph Optimization} (PGO). However, these methods require the initial pose errors to be small to find the correct object matching in the first step and then iterate many times, or use the PGO to refine the result. In this paper, we propose to learn pose-agnostic neighborhood graph features to accurately find the correct matching and then use PGO to refine it.
\textbf{Communication delay} can cause that data arriving at the ego agent is outdated. To compensate for this delay, \cite{Yu2023v2xseq,yu2023ffnet} add the first-order derivatives of the BEV feature map to the original feature map to obtain the updated feature map that is aligned to the ego timestamp; \cite{syncnet} employs the LSTM~\cite{lstm} structure on the BEV feature maps to adjust the features to be aligned. Differently, \cite{streamlts} embeds the observation timestamps as well as the temporal context from historical frames into the object queries so that they are time- and motion-aware and capable of predicting the status at any given future timestamp, achieving the time-delay compensation. Different to \cite{Yu2023v2xseq,yu2023ffnet,syncnet} that compensate the feature maps, \cite{streamlts} leverages the point-wise timestamp of the point clouds to learn the accurate temporal context, hence is also able to cope with \textbf{sensor asynchrony} issues. Therefore, we adopt the query-based strategy for time-related compensation.

\subsection{3D Backbones}
To extract point cloud features, most previous works~\cite{Xu2022opv2v,Wang2020v2vnet,Yin2023v2vformer_plus,coalign,Yu2023v2xseq,v2vam,fpvrcnn} for COOD either use dense 3D~\cite{voxelnet,xia2021soe}, dense 2D~\cite{pointpillar,pixor} or semi-sparse~\cite{pvrcnn,zheng2020ciassd,second,Xia_2023_ICCV} backbones that first encode the point cloud with 3D sparse convolution and then apply dense 2D convolutions for further processing. In contrast, \cite{Yuan_gevbev2023,streamlts} employ the MinkUnet~\cite{minkunet} to extract point cloud features. Despite its efficiency, the MinkUnet backbone struggles with long-range detection, occlusions and large objects, because of CFM and ICF problems. \cite{fan2022fully} proposed center voting for each scanned object points to solve CFM issue and \cite{voxelnext} reveals that further down-sampling the voxel features to lower resolutions can increase the receptive field and refine the large object detection. Instead of shifting the feature locations as \cite{fan2022fully} to accurately locate object queries or adding additional layers as \cite{voxelnext}, we build a backbone that solves both CFM and ICF issues by modifying MinkUnet with CEC layers.

\section{Method}

\subsection{Task formalization}
In this work, we solve the LiDAR-based cooperative object detection task considering  unavoidable challenges in reality: communication delay, sensor asynchrony, and localization errors. For this scenario, we assume there are $N$ IAs, including one ego agent $A_1$ and $N-1$ cooperative agents. These agents are sharing CPMs with each other in a broadcasting manner, as shown in \cref{fig:sparse_issues}b).  
Specifically, in the $i$-th frame, $A_1$ receives the CPMs $M=\{(m_j, t_j)| j\in \{1, \dots, N-1\}\}$ from the cooperative agents, where $m_j$ and $t_j$ are the CPM and the corresponding timestamp from the $j$-th agent. We aim to optimize the object detection results based on the point clouds captured so far ($t\leq t_i$) at $A_1$ and the received messages $M$. We model and evaluate both the COOD and TA-COOD tasks from the perspective of an ego agent. The key difference between COOD and TA-COOD lies in the  GT BBoxes.  As illustrated in \cref{fig:pipeline}, TAM, COOD GT BBoxes (dashed lines) align with the sequentially captured LiDAR points, whereas TA-COOD GT BBoxes (solid lines) are referenced to a global timestamp $t_g$. This distinction requires the model to accurately predict BBox locations at $t_g$. While COOD relies solely on point cloud geometry for location prediction, TA-COOD leverages temporal context to enhance accuracy at $t_g$ and also enables the model to handle location errors due to communication latency.

\subsection{Framework overview}
We build an efficient fully sparse framework \textit{SparseAlign}, as shown in \cref{fig:pipeline}, to process sequential point clouds from multiple agents. All agents share the same network weights. The input at time $t_i$ is the point cloud of the ego agent $PC_{t_i}$ and of the cooperative agents $PC_{t_j}$. Note that $t_j<t_i$ when communication latency exists. All point clouds are processed by the backbone SUNet to extract the point cloud features. The Region of Interest (RoI) module then selects the most interesting features as the object queries. For simplicity, we define object queries as $Q^* = \{(F^*_i, x^*_i, y^*_i) | i\in \{1, \dots, N^*\}$, where $*$ represents the query set name, $F$ is the features, and $x, y$ are the coordinates of the corresponding $i$-th query. 
With selected queries, the TempAlign Module (TAM) can efficiently align the query features to the globally aligned timestamp $t$ so that errors introduced by communication delay and sensor asynchrony are compensated. The updated query features and the detections from the \textbf{L}ocal \textbf{Q}uery \textbf{Det}ection head (LQDet) are then shared as CPMs.  The PoseAlign Module (PAM) then uses the detected BBoxes to correct the relative poses between the cooperative and the ego agent and the SpatialAlign Module (SAM) fuses the received cooperative features into the ego coordinate system based on the corrected transformation $T_c^e$ from cooperative to ego coordinates. 

\begin{figure}[t]
    \centering
    \includegraphics[width=\linewidth]{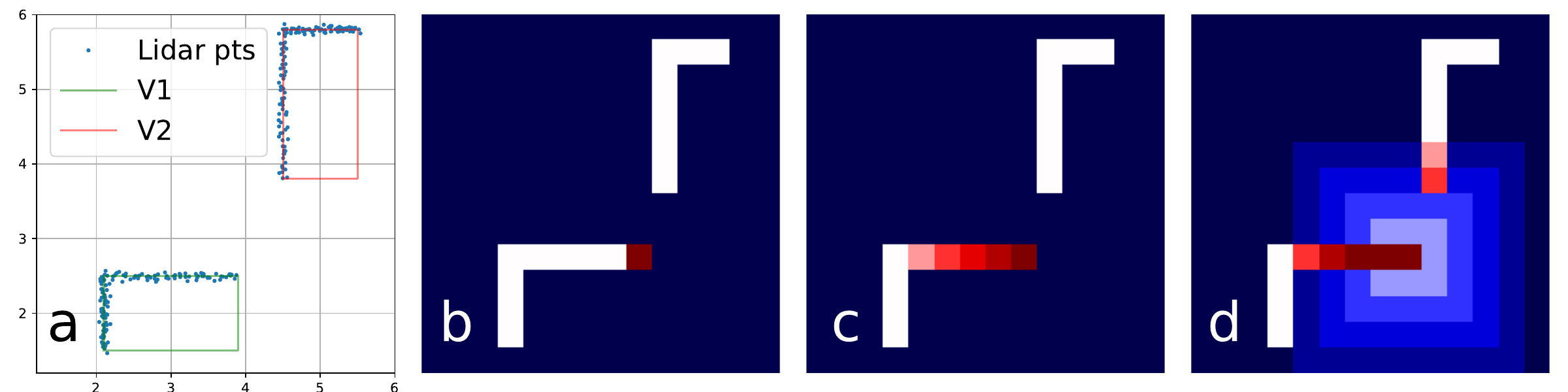}
    \caption{Receptive field (RF) after 4 sparse convolutions (convs). a. LiDAR points; b. Spare pixels (white) with a conv center point (red); c. RF coverage of 4 normal sparse convs (red); d. RF coverage of 4 CECs (red+light blue, darkest blue is background).}
    \label{fig:icf_cfm}
    \vspace{-10pt}
\end{figure}
\subsection{Backbone: Sparse UNet (SUNet)}
Processing data from multiple IAs for cooperative perception is computationally demanding. To improve efficiency, we use a fully sparse 3D UNet, similar to MinkUnet\cite{minkunet}, to construct the SUNet (\cref{fig:pipeline} bottom left). 
However, the fully sparse convolutions used in such networks suffer from the ICF issue. As shown in \cref{fig:icf_cfm}, the receptive field (RF) of (c) standard sparse convolutions only covers LiDAR points from a single vehicle. In contrast, the RF of (d) the CEC layers extends to the points of a neighboring vehicle, enabling learning within a larger spatial context.
Therefore, we use 3D CECs in the sparse convolution (SConv) block 3 and 4. Correspondingly, we contract the coordinates in the transposed sparse convolution (TConv) block 3 and 2. In this way, the output of the TConv blocks $p4$ and $p2$ can match the coordinates in stride $x2$ and $x4$ for the concatenation. In addition, it learns better global features without voxel number soaring in the subsequent layers and maintains efficiency. 
Against the CFM issue, we use 2D CECs on the 2D BEV sparse features (\cref{fig:pipeline} {SUNet:Coor. Expand.}) to expand the coordinates to ensure their coverage over the object centers.

\subsection{Alignment modules}

\textbf{TempAlign Module (TAM).}
The TAM aims to learn the temporal context relative to the previous frames to achieve two goals: improve object detection performance by looking for clues in the history and compensate for the object displacement error introduced by asynchronous scanning time (\cref{fig:pipeline}, TAM). We achieve the second goal by predicting the BBoxes at the globally aligned timestamp so that the BBoxes shared for PAM are synchronized. Then PAM only needs to deal with the object displacement caused by localization errors. Technically, the TAM follows \cite{streamlts}, for efficiency, by removing the global cross-attention module which learns the global BEV features.  We only use a memory queue and a hybrid attention module. The memory queue stores the top $k=256$ object queries from previous $L=4$ data frames so that the attention module can allow the interaction between object queries in the current frame and the historical frames. In this way, the object queries learn the temporal context from the sequential data and are able to make predictions for the future state. 

\noindent
\textbf{PoseAlign Module (PAM).}
Given the detected BBoxes $\mathbf{B}_i$ and $\mathbf{B}_j$, respectively, from the perspective of the agent $A_i$ and $A_j$ in the same scene, humans can easily find the correspondences between $\mathbf{B}_i$ and $\mathbf{B}_j$ based on the geometric relationships. Practically, we model the features of these geometric relationships by embedding the neighborhood geometry of each BBox into deep features and then match these features of $\mathbf{B}_i$ and $\mathbf{B}_j$ with the Hungarian algorithm. For each BBox $b$ in a BBox set $\mathbf{B}$, we select the nearest $k=8$ neighbors to embed its neighborhood features. As shown in \cref{fig:pipeline}, PAM: \textit{1.Graph Feature}, the orientation angles of the BBox are $\alpha$ and $\beta$ for $b$ and its neighbor, respectively.  We embed the relative orientation with $\nu_a=[\sin(\beta - \alpha); \cos(\beta - \alpha)]$\footnote{$[\cdot ; \cdot]$: concatenation}. Similarly, the relative edge direction is embedded as $\epsilon_a=[\sin(\theta - \alpha); \cos(\theta - \alpha)]$. In addition, the dimension $\nu_{dim}=[l, w, h]$ of the neighbor BBox, and the Euclidean distance between $b$ and its neighbors $\epsilon_d=d$, are also embedded into the feature. 
Note that $\epsilon_d, \epsilon_a, \nu_a$ are all relative features between two BBoxes that are pose-agnostic, leading to robust pose alignment without dependency on the magnitude of initial pose errors.
These features are then concatenated and embedded into high-dimensional features by a $linear$ layer with ReLU activation ($\mathbb{R}^{5}\rightarrow \mathbb{R}^{d_{emb}}$):
\begin{equation}\label{eq:linear}
    f_{nbr} = linear([\epsilon_d; \epsilon_a; \nu_a; \nu_{dim}])
\end{equation}

Assembling the features relative to all neighbors, we summarize the neighborhood feature $F_{nbr}\in \mathbb{R}^{k\times d_{emb}}$ by a multi-head self-attention module $attn$ followed by a $linear$ layer operated on the $mean$ and $max$ of the $attn$ outcome, resulting in the feature $F_{nbr}\in \mathbb{R}^{d_{emb}}$:
\begin{align}
    F_{nbr} &= attn(F_{nbr})\\
    F_{nbr} &= linear([mean(F_{nbr}); max(F_{nbr})])
\end{align}

By calculating the Euclidean distance between the learned neighborhood feature $F_{nbr}$ of the BBox in $\mathbf{B}_i$ and $\mathbf{B}_j$, we construct the cost matrix for the linear sum assignment to find the best \textit{Match} (\cref{fig:pipeline}, PAM: \textit{2.Match}) between these two BBox sets. However, an object in $\mathbf{B}_i$ does not always have a projection in $\mathbf{B}_j$ and vice versa. Thus, we reject the wrong matches that have large distances.
Based on the PAM: \textit{2.Match} result, the relative pose transformation between $A_i$ and $A_j$ can be recovered, completing the PAM: \textit{3.Align} step. Additionally, we use pose graph optimization to refine the alignment by leveraging loop closure among multiple cooperative agent poses following \cite{coalign}.

\noindent
\textbf{SpatialAlign Module (SAM).}
To align the cooperative query features $Q^c$ with the ego query features $Q^e$, there are two issues to be resolved. First, the features $F^c$ are learned in the cooperative coordinate system and must be adapted to the ego coordinate system. We achieve this using a multilayer perceptron ($MLP$) as described in \cref{eq:rot_adapt} where $F^R\in \mathbb{R}^{N\times 9}$ is the flattened and repeated rotation matrix $R\in \mathbb{R}^{3\times 3}$. 
\begin{equation}\label{eq:rot_adapt}
    F^c = MLP([F^c; F^R])
\end{equation}
Second, after rotation with $R$, the coordinates of $Q^c$ (\cref{fig:pipeline}, SAM, yellow points) do not perfectly align to the discretized grid in the ego coordinate system. Thus, we merge $Q^c$ into the nearest points of this grid. The locations of the fused output object query points $Q^{ts}$ are shown with gray points. For each point in $Q^{ts}$, we summarize the features of its $k=8$ nearest neighbors (\cref{fig:pipeline}, SAM: black arrows) in query set $Q^{\cup}=Q^c \bigcup Q^e$ to obtain the fused features $F^{ts}$. Mathematically, it reads as
\begin{align}
    F_{ij} &= mlp([F^{\cup}_j; linear([x^{ts}_i-x^{\cup}_j, y^{ts}_i-y^{\cup}_j])])\label{eq:sa_fusion1}\\
    F^{ts}_i &= \{F_{ij} | j\in \{1, \dots, k\}\}\label{eq:sa_fusion2}\\
    F^{ts}_i &= max_j(F^{ts}_i) + mean_j(F^{ts}_i) \label{eq:sa_fusion3}
\end{align}
where $i$ is the query index of $Q^{ts}$ and $j$ is the neighbor index of the $i$-th query.

\subsection{Detection Heads}
\begin{figure}
    \centering
    \includegraphics[width=0.95\linewidth]{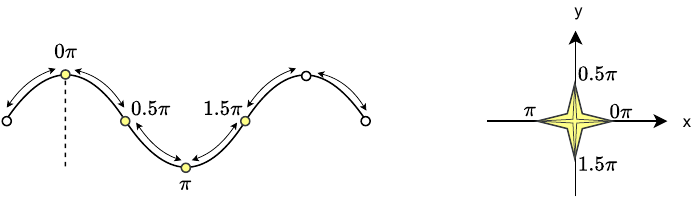}
    \caption{CompassRose  encoding}
    \label{fig:compass_rose}
\end{figure}

To train \textit{SparseAlign}, we follow \cite{streamlts} to attach a center-based object detection head to each of the RoI, TAM and SAM to learn query features. These heads are notated as \textit{RoIDet}, \textit{LQDet} and \textit{GQDet}, respectively. They share the same target-encoding method. We use \textit{Focal} loss to classify fore- and background object queries; all query points inside the ground-truth BBoxes are positive, and those outside are negative. \textit{Smooth L1} loss is used for regressing $[dx,dy,dz,l,w,h,\mathcal{B}_{dir},\mathcal{B}_{scr}]$, where $l, w, h$ are the dimensions of BBoxes and $dx,dy,dz$ are coordinate offsets between the query points and the center of ground-truth BBoxes. $\mathcal{B}_{dir},\mathcal{B}_{scr}$ are the novel \textit{CompassRose} encoding (\cref{fig:compass_rose}) of the ground truth orientation $r_g$ with respect to four anchor angles $\mathbf{r}_a = [0, 0.5\pi, \pi, 1.5\pi]$. It reads as
\begin{align}\label{eq:compass_rose} 
    \mathcal{B}_\text{dir} &= [\cos{r_g} -  \cos{\mathbf{r}_a},
    \sin{r_g} - \sin{\mathbf{r}_a}]\\
    \mathcal{B}_\text{scr} &= [s_{0}, s_{0.5\pi}, s_{\pi}, s_{1.5\pi}]\\
    &=\mathbf{1} - \arccos{(\cos{\mathbf{r}_a}\cdot \cos{r_g} + \sin{\mathbf{r}_a}\cdot \sin{r_g})} / \pi \nonumber
\end{align}

Take the encoded $sine$ wave (\cref{fig:compass_rose} left) as an example, the four anchor angles (yellow points) ensure that at least one anchor can reach the target angle monotonically, simplifying the regression process.

\subsection{Data augmentation}
In addition to geometric augmentation, which includes randomly flipping the input point cloud along the x- or y-axis and rotating it around the z-axis, we apply Free Space Augmentation (FSA) \cite{Yuan_gevbev2023}. FSA enhances the point cloud by adding auxiliary points along LiDAR scanning rays, representing empty spaces where no objects are present. This augmentation helps mitigate the ICF problem in distant regions, where observation points are extremely sparse. For more details, refer to \textit{supplementary B}.

\begin{table*}[t]
\centering
\begin{minipage}{.48\textwidth}
\resizebox{\textwidth}{!}{ 
\begin{tabular}{l|c|c|c}
\toprule
{Method} & BW(Mb) $\downarrow$ & {AP0.5} $\uparrow$& {AP0.7}$\uparrow$\\ 
 \hline
Fcooper\cite{fcooper}  & 72.08    & 0.887 &  0.790   \\ 
V2VNet\cite{Wang2020v2vnet} & 72.08   & 0.917  &  0.822  \\ 
FPVRCNN\cite{fpvrcnn}  & 0.24   & 0.873  &  0.820  \\ 
OPV2V\cite{Xu2022opv2v} & 126.8   & 0.905  &  0.815 \\ 
CoAlign\cite{coalign}  & 72.08   & 0.902  &  0.833  \\  
CoBEVT\cite{xu2022cobevt}  & 72.08   & 0.913  &  0.861  \\ 
V2VAM\cite{v2vam} & 72.08   & 0.916  &  0.850  \\ 
V2VFormer++\cite{v2vformer_plus}$^{L+C}$ & 72.08  & \textbf{0.935} & \textbf{0.895} \\
SparseAlign (ours)  & \textbf{$<$ 1.3}   & \underline{0.930}  &  \underline{0.892}  
 \\ \bottomrule
\end{tabular}
}
\caption{COOD Average Precision on OPV2V dataset.}\footnotemark
\label{tab:sota_opv2v}
\end{minipage}
\hfill
\begin{minipage}{.45\textwidth}
\resizebox{\textwidth}{!}{ 
\begin{tabular}{l|c|c|c}
\toprule
{Method} & BW(Mb) $\downarrow$ & {AP0.5} $\uparrow$& {AP0.7}$\uparrow$\\ 
 \hline
Fcooper\cite{fcooper} & 48.8    & 0.734  &  0.559 \\ 
V2VNet\cite{Wang2020v2vnet} & 48.8    & 0.664  &  0.402 \\ 
FPVRCNN\cite{fpvrcnn} & 0.24   & 0.665  &  0.505 \\ 
V2XViT\cite{Xu2022v2xvit}& 48.8    & 0.704  &  0.531 \\ 
DiscoNet\cite{xu2022cobevt} & 48.8   & 0.736  &  0.583 \\ 
OPV2V\cite{Xu2022opv2v} & 97.6   & 0.733  &  0.553 \\ 
CoAlign\cite{coalign} & 48.8   & 0.746  &  0.604 \\ 
DI-V2X\cite{div2x} & 48.8 & \underline{0.788} & \underline{0.662}  \\
SparseAlign (ours) & \textbf{$<$ 1.3 }   & \textbf{0.845} &  \textbf{0.685}  
 \\ \bottomrule
\end{tabular}
}
\caption{COOD Average Precision on DairV2X dataset.}\label{tab:sota_dairv2x}
\end{minipage}
\end{table*}

\begin{table}[tb]
\centering
\resizebox{\linewidth}{!}{ 
\begin{tabular}{l|cc|cc}
\toprule
\multirow{2}{*}{Method} & \multicolumn{2}{c|}{OPV2Vt} & \multicolumn{2}{c}{DairV2Xt}\\ 
 &  \multicolumn{1}{c}{AP0.5$\uparrow$}    &  AP0.7$\uparrow$   & \multicolumn{1}{c}{AP0.5$\uparrow$} &  AP0.7$\uparrow$\\ 
 \hline
Fcooper\cite{fcooper}+SA& 0.763 & 0.553 & 0.597 & 0.282\\ 
FPVRCNN\cite{zheng2020ciassd}+SA& 0.640 & 0.474 & 0.598& 0.307 \\ 
OPV2V\cite{Xu2022opv2v}+SA& \underline{0.881} &0.787 & 0.702& 0.366 \\
StreamLTS\cite{streamlts} & 0.853 & 0.721  & 0.640  & 0.404\\
StreamLTS\cite{streamlts}+SA& 0.850 & \underline{0.790} & \underline{0.757} & \underline{0.497}\\
SparseAlign (full)  & \textbf{0.893} &  \textbf{0.818} & \textbf{0.796} & \textbf{0.548}    
 \\ \bottomrule
\end{tabular}

}
\caption{TA-COOD Average Precision on OPV2Vt and DairV2Xt}\label{tab:ta_cood}
\end{table}

\section{Experiment}

\textbf{Datasets.} We evaluate \textit{SparseAlign} on OPV2V\cite{Xu2022opv2v} and DairV2X\cite{Yu2022dairv2x} datasets for COOD and their variant datasets OPV2Vt\cite{streamlts} and DairV2Xt\cite{streamlts} for TA-COOD. OPV2V is a synthetic dataset created in different driving scenarios of CARLA~\cite{carla} simulator. Following \cite{Xu2022opv2v}, the detection range is set to $x\in [-140, 140]m, y\in [-40, 40]m$. DairV2X is a real dataset containing one vehicle- and one infrastructure-side sensor suit. The detection range is set to $x\in [-100, 100]m, y\in [-40, 40]m$. OPV2Vt is generated by interpolating OPV2V data frames into sub-frames and replaying them to mimic the sensor asynchrony and the rolling shutter problem of LiDARs. The GT BBoxes of this dataset are aligned to the global timestamp, the LiDAR scan ending time of each frame. DairV2Xt follows the same idea. The detection range of OPV2Vt and DairV2Xt is set the same as OPV2V and DairV2X.

\noindent
\textbf{Implementation details.} All models are trained on a single GTX-4090 and an Intel i7-8700 CPU for $50$ epochs (about 35 hours) with the Adam optimizer and batch size two. The learning rate is set to $2e-4$ with a warm-up stage of $4000$ iterations. The random rotation angle is set to the range $[-90^\circ, 90^\circ]$.
For models incorporating the TAM, we load four sequential frames of data during training and calculate losses solely for the last frame. To reduce GPU memory consumption, we enable gradient calculations only for data from a single ego agent and one cooperative agent. For details on efficient gradient scheduling, refer to \textit{supplementary D} . To ensure that the model generalizes well to the significant time latency introduced by communication, we load data from cooperative vehicles with a random latency ranging from 0 to 200 ms relative to the ego vehicles. 
 
\footnotetext{The results reported in this table use frame-wise local score sorting for AP, same to the official benchmark. Other results use global sorting.}
\begin{table*}[ht]
\centering
\begin{minipage}{0.55\textwidth}
\centering
\resizebox{\textwidth}{!}{ 
\begin{tabular}{c|c|c|c|cc|cc}
\toprule
CEC & CEC  & TAM & SAM & \multicolumn{2}{c|}{OPV2V} & \multicolumn{2}{c}{DairV2X} \\ 
2D & 3D &  & &\multicolumn{1}{c}{ AP0.5}    &  AP0.7&\multicolumn{1}{c}{ AP0.5}    &  AP0.7 \\ \hline 
\checkmark &\checkmark &\checkmark &           &0.847 & 0.756  & 0.671 & 0.528\\
\checkmark &\checkmark &\checkmark & QST       &0.888 & 0.869  & 0.749 & 0.622\\
           &           &           &\checkmark &0.924  & 0.885  & 0.773 & 0.638\\ 
\checkmark &           &           &\checkmark &0.940  & 0.901  & 0.796 & 0.662\\ 
\checkmark &\checkmark &           &\checkmark &0.950  & 0.914  & 0.813 & 0.682\\
\checkmark &\checkmark &\checkmark &\checkmark &\textbf{0.951}  & \textbf{0.929}  & \textbf{0.845} & \textbf{0.685}\\
\bottomrule
\end{tabular}
}
\caption{Ablation study on 2D and 3D CEC layers, TAM and SAM. QST: use QUEST~\cite{fan2023quest} spatial fusion instead of SAM.}\label{tab:ablation}
\end{minipage}%
\hfill
\begin{minipage}{0.4\textwidth}
\centering
\resizebox{\textwidth}{!}{ 
\begin{tabular}{c|cc|cc}
\toprule
Dir. & \multicolumn{2}{c|}{OPV2V} & \multicolumn{2}{c}{DairV2X} \\ 
Enc.&\multicolumn{1}{c}{ AP0.5} & AP0.7 & \multicolumn{1}{c}{ AP0.5} & AP0.7 \\ \hline 
\textit{gt-angle} & 0.871 & 0.768 & 0.620 & 0.427 \\ 
\textit{second} & 0.908 & 0.874 & 0.743 & 0.590 \\ 
\textit{sin\_cos} & 0.915 & 0.881 & 0.748 & 0.603 \\ 
\textit{compass} & \textbf{0.924} & \textbf{0.885} & \textbf{0.773} & \textbf{0.638} \\ 
\bottomrule
\end{tabular}
}
\caption{Comparison on BBox angle encoding methods: offset to ground-truth angle (\textit{gt-angle}), second-style (\textit{second}), sine-cosine offsets (\textit{sin\_cos}) and compass rose (\textit{compass}).}\label{tab:ablation2}
\end{minipage}
\end{table*}

\begin{table*}
\centering
\resizebox{0.95\textwidth}{!}{ 
\begin{tabular}{c|cc|cc|cc|cc|cc|cc}
\toprule
 Dataset & \multicolumn{6}{c|}{OPV2Vt}  & \multicolumn{6}{c}{DairV2Xt}\\ \cline{1-13} 
Latency & \multicolumn{2}{c}{0ms} & \multicolumn{2}{c}{100ms} & \multicolumn{2}{c|}{200ms}& \multicolumn{2}{c}{0ms} & \multicolumn{2}{c}{100ms} & \multicolumn{2}{c}{200ms}\\\cline{1-13} 

AP threshold & \multicolumn{1}{c}{0.5\quad}    &  0.7\quad   & \multicolumn{1}{c}{0.5} &  0.7 & \multicolumn{1}{c}{0.5} &  0.7& \multicolumn{1}{c}{0.5}    &  0.7   & \multicolumn{1}{c}{0.5} &  0.7 & \multicolumn{1}{c}{0.5} &  0.7\\ \hline
SteamLTS~\cite{streamlts}
& \multicolumn{1}{c}{0.853} & 0.721 & \multicolumn{1}{c}{0.816} & 0.680 & 0.787 & 0.647
& \multicolumn{1}{c}{0.642} & 0.404 & \multicolumn{1}{c}{0.613} & 0.379 & 0.590 & 0.364\\ 
SA no TAM
&\multicolumn{1}{c}{\underline{0.890}} & 0.802 & \multicolumn{1}{c}{0.856} & 0.471 & \multicolumn{1}{c}{0.631} & 0.209 
&\multicolumn{1}{c}{0.720} & 0.457 & 0.672 & 0.393 & 0.635 & 0.350\\  
SA no PT
& \multicolumn{1}{c}{0.835} & 0.615 & \multicolumn{1}{c}{0.831} & 0.625 & 0.815 & 0.581
& \multicolumn{1}{c}{0.760} & 0.469 & \multicolumn{1}{c}{0.737} & 0.459 & 0.722 & 0.449\\ 
SA no TL
& \multicolumn{1}{c}{0.885} & \textbf{0.843} & \multicolumn{1}{c}{0.862} & 0.619 & 0.723 & 0.279
& \multicolumn{1}{c}{\underline{0.793}} & \underline{0.534} & \multicolumn{1}{c}{0.749} & 0.474 & 0.698 & 0.423\\  
SA no FSA
& \multicolumn{1}{c}{0.881} & 0.816 & \multicolumn{1}{c}{\underline{0.875}} & \textbf{0.795} & \underline{0.857} & \underline{0.762}
& \multicolumn{1}{c}{0.769} & 0.508 & \multicolumn{1}{c}{\underline{0.755}} & \underline{0.502} & \underline{0.740} & \underline{0.496}\\ 
SparseAlign (SA)
& \multicolumn{1}{c}{\textbf{0.893}} & \underline{0.818} & \multicolumn{1}{c}{\textbf{0.885}} & \textbf{0.795} & \textbf{0.867} & \textbf{0.765}
& \multicolumn{1}{c}{\textbf{0.796}} & \textbf{0.548} & \multicolumn{1}{c}{\textbf{0.786}} & \textbf{0.543} & \textbf{0.772} & \textbf{0.532}\\ 
\bottomrule
\end{tabular}
}
\caption{AP of TA-COOD with communication latency. PT: Point-wise Timestamps, TL: Train Latency, FSA: Free Space Augmentation. }\label{tab:ablation_latency}
\end{table*}

\section{Results}
\subsection{Comparison to the state-of-the-art}
\cref{tab:sota_opv2v} presents the COOD results on the OPV2V dataset. All methods use only LiDAR data, except for \textit{V2VFormer++} (denoted by L+C), which incorporates both LiDAR and camera data. Compared to the state-of-the-art, our \textit{SparseAlign} achieves the best performance in LiDAR-based COOD, approaching the results of \textit{V2VFormer++}, which benefits from LiDAR-Camera fusion. \cref{tab:sota_dairv2x} shows the results on the DairV2X dataset. Here, \textit{SparseAlign} outperforms other methods by a large margin, achieving a 2.3\% improvement in AP at the IoU threshold of 0.7 and a 5.7\% improvement at the IoU threshold of 0.5. This is mainly due to the enhanced temporal and global reasoning ability of our model for detecting distant and large objects.
Compared to other methods that rely on  dense BEV feature maps, \textit{SparseAlign}  consumes the least communication bandwidth (BW) of less than 1.3MB (without compression) thanks to the efficient query-based operations.

For the results of the more challenging TA-COOD task shown in \cref{tab:ta_cood}, our \textit{SparseAlign} outperforms \textit{StreamLTS} by a great margin. By applying our three alignment modules (\textit{StreamLTS+SA}), the performance of StreamLTS is also improved. 
In addition, we compare our framework to \textit{Fcooper}, \textit{FPVRCNN} and \textit{OPV2V} method by attaching the three alignment modules to them to achieve TA-COOD. All show worse performance than the full \textit{SparseAlign} because of worse 3D backbone or the RoI selection.

The qualitative results and the additional comparison to methods, such as \textit{TransIFF}\cite{TransIFF}, on the 3D AP metric are given in the \textit{supplementary} A and E, respectively.

\subsection{Ablation study}
We conduct an ablation study on the 3D backbone, TAM, and SAM (PAM is discussed separately in \cref{sec:loc_err}). Among these components, the spatial fusion module plays a crucial role in multi-agent feature fusion. Without {SAM}, the module performs the worst, as evidenced by the first row of \cref{tab:ablation}. When using QUEST \cite{fan2023quest} spatial fusion, which aligns only rotation (\cref{eq:rot_adapt}) without correcting spatial displacement (\cref{eq:sa_fusion1,eq:sa_fusion2,eq:sa_fusion3}), the AP scores remain significantly lower than those obtained with {SAM}. Incorporating {SAM} improves performance, and we further enhance it by gradually replacing standard sparse convolutional layers with CEC layers. Finally, {TAM} further boosts AP scores by leveraging temporal information from historical frames.

In \cref{tab:ablation2}, we compare CompassRose encoding with three methods: directly regress the ground-truth angle (\textit{gt-angle}), second-style~\cite{second} encoding (\textit{second}), and the sine-cosine encoding (\textit{sin\_cos}). For simplicity, we use the configurations in the third row of \cref{tab:ablation} for the comparison. The results show that CompassRose performs the best.

\subsection{Communication latency and sensor asynchrony}
\cref{tab:ablation_latency} shows the results of TA-COOD on the OPV2Vt and the DairV2Xt dataset. Our proposed \textit{SparseAlign} demonstrates a significant improvement over the baseline framework, \textit{StreamLTS}~\cite{streamlts}. Specifically, the AP at an IoU threshold of 0.7 increased 9.7\% on the OPV2Vt dataset and 14.4\% on the DairV2Xt dataset. As the communication latency increases to 200ms, the AP at IoU 0.5 of our \textit{SparseAlign} only dropped 2.6\% while the baseline work has lost 6.6\% of its accuracy. This reveals that our framework is more robust against communication latency. To investigate modules or configurations that are making an effect on this robustness, we conducted an ablation study by removing the TAM, Point-wise Timestamps (TP), Training Latency (TL, communication latency during training) and Free Space Augmentation (FSA), respectively. 

Without the TAM, the performance noticeably declines as latency increases, despite achieving comparable results to the full \textit{SpareAlign} model at 0 ms latency. A similar effect is observed when the full SA model is trained without exposure to data with communication latency (SA without TL). Remarkably, without TL, the model outperforms the full \textit{SpareAlign} model in terms of AP at an IoU threshold of 0.7 on the OPV2Vt dataset (0.843 against 0.818). This is because the model focuses more on predicting BBoxes within the same frame, where only minimal latency is introduced by asynchronous sensors. 

Without PT, the model performs much worse than the full \textit{SpareAlign}, showing the importance of fine-grained point-wise timestamps in the point clouds for the model to compensate for the errors introduced by sensor asynchrony and capture the accurate temporal context for prediction. Finally, the ablation study on the FSA module also shows a positive influence on the performance. For example, AP at the IoU of 0.7 increased by 4\% on DairV2Xt by utilizing the FSA (See \textit{supplementary} D for visual comparison).

\subsection{Localization errors}\label{sec:loc_err}

\begin{figure}[t]
    \centering
    \includegraphics[width=0.9\linewidth]{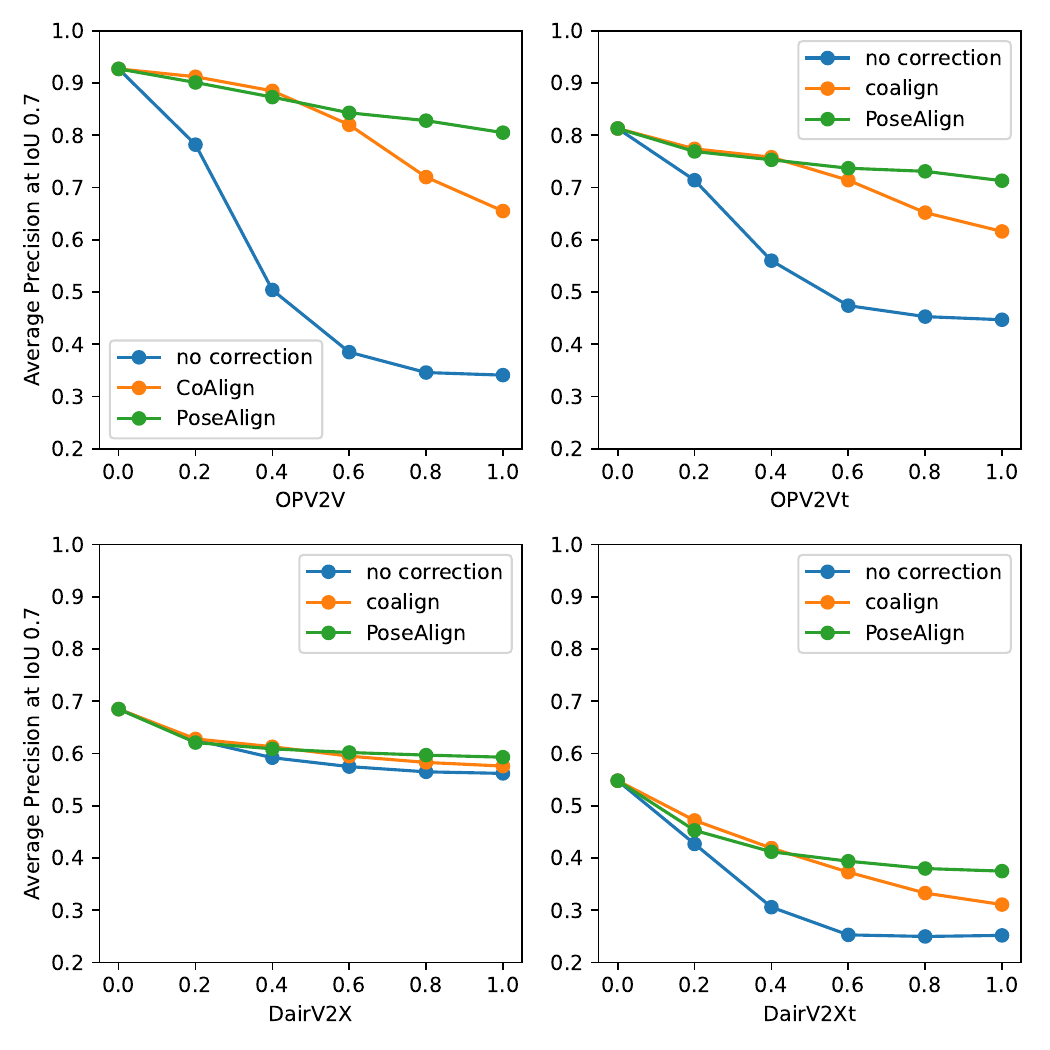}
    \caption{AP at IoU threshold of 0.7 with translation errors ranging from 0m to 1m along x- and y-axis, and rotation errors from $0^\circ$ to $1.0^\circ$ (horizontal axis) for the different datasets.}
    \label{fig:loc_err}
\end{figure}

\cref{fig:loc_err} shows the performance of \textit{SpareAlign} against different localization errors. We add random errors to the poses of both ego ($T_e$) and cooperative ($T_c$) vehicles. The translation errors $x_\epsilon, y_\epsilon$ along the $x$- and $y$-axis and the rotation error $r_\epsilon$ around the $z$-axis are all assumed to be normally distributed with $\mathcal{N}(0, 1) \cdot \epsilon$, where $\epsilon\in [0, 1]$ is the error scaling factor. We report the results by gradually increasing $\epsilon$ with a step size of $0.2$. Note that the translation and rotation errors exist in the poses of both the ego and the cooperative vehicle; the resulting relative translation error of the transformation $T_c^e=inverse(T_e)\cdot T_c$ from cooperative to ego coordinate system will be amplified as the rotation error increases. Our \textit{PAM} is specially designed to mitigate the influence of large relative transformation errors. 

Our proposed \textit{PAM} and the baseline method \textit{CoAlign}~\cite{coalign}  both significantly reduce the impact of pose errors on AP performance compared to the configuration lacking any correction module. 
In addition, our method demonstrates greater robustness against large errors. For instance, on the OPV2V dataset, the AP for \textit{CoAlign} decreased by approximately $27\%$ at the error of $1.0m, 1.0m, 1.0^\circ$, whereas {PAM} experienced only a $12\%$ drop.
This validates the efficacy of our design, which relies on the pose-agnostic relative neighborhood geometries to match the detected BBoxes of the ego and the cooperative IAs. Note that the DairV2X dataset exhibits smaller changes in AP as localization errors increase, because the pose parameters are not well calibrated. In contrast, \cite{streamlts} refined the poses of DairV2X for generating the DairV2Xt dataset,  making the better-calibrated DairV2Xt more sensitive to newly introduced errors.
\begin{figure}
\centering
    \includegraphics[width=0.9\linewidth]{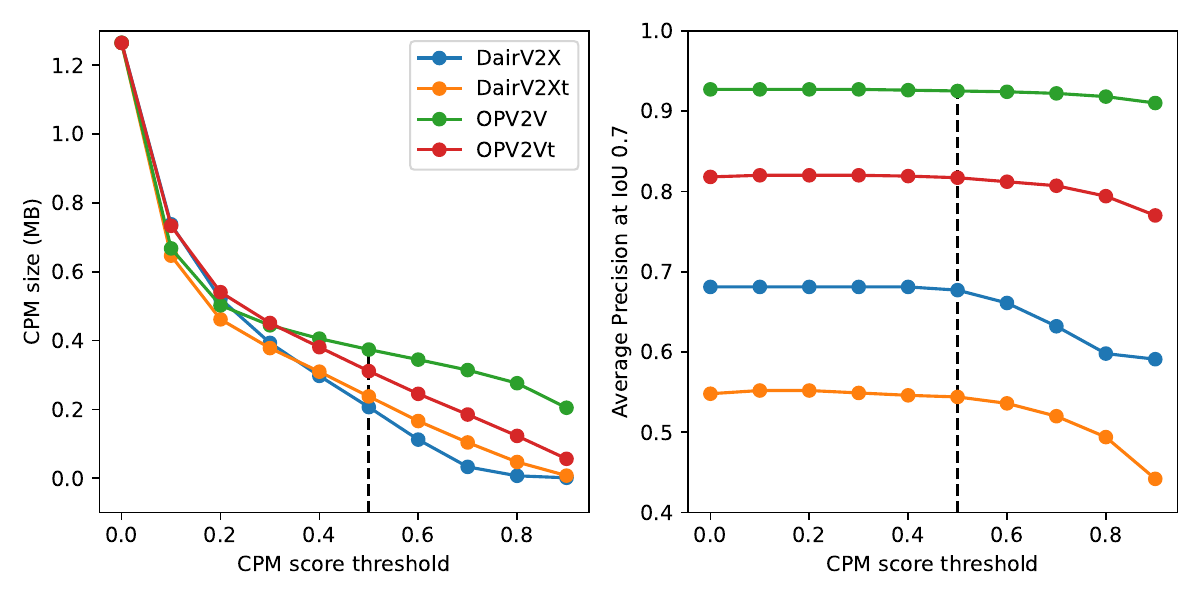}
    \caption{CPM sizes with different information selection scores. Object query features with detection scores larger than the threshold (x-axis) are shared as CPMs.}
    \label{fig:cpm_sizes}
\end{figure}
\subsection{CPM sizes}
Our framework significantly reduces communication bandwidth without relying on computationally intensive dense feature maps, unlike state-of-the-art methods. This efficiency is achieved by selecting only the top $K=1024$ object queries for further processing. In addition to this intrinsic reduction, the size of the CPMs can be further minimized through score-based selection. Specifically, we employ the query-based detection scores from the {LQDet} head to identify and share only the most critical queries. As \cref{fig:cpm_sizes} shows, using a CPM score threshold below 0.5 does not lead to a noticeable degradation in performance across all datasets. Remarkably, at a threshold score of 0.5, the average size of the CPMs is reduced to less than 400 KB on all datasets. The final CPM sizes after selection are influenced by the driving scenes. Specifically, scenes with more vehicles tend to have larger CPM sizes. Consequently, the OPV2V and OPV2Vt datasets, which contain more vehicles in their scenes, result in larger CPM sizes compared to the DairV2X and DairV2Xt datasets.

\section{Conclusion}

In this paper, we proposed \textit{SparseAlign}, a fully sparse framework for cooperative object detection (COOD). With our novel 3D backbone network {SUNet}, the {CompassRose} encoding of detection head, the temporal and spatial fusion module TAM and SAM, \textit{SparseAlign} significantly outperforms state-of-the-art methods on the OPV2V and DairV2X datasets for COOD, as well as on their variants OPV2Vt and DairV2Xt for time-aligned COOD (TA-COOD), achieving this with very low communication bandwidth. Additionally, our proposed pose alignment module PAM demonstrates improved robustness to large localization errors compared to the baseline method CoAlign. Nevertheless, when the pose errors are large, a drop of about 10\% in APs persists even with pose correction. In future work, we therefore plan to explore more robust methods, \eg~leveraging sub-graph detection, to further enhance the pose alignment accuracy. Additionally, more application scenarios, such as roadside cooperative detection benchmarks HoloVIC\cite{holovic}, RCooper\cite{hao2024rcooper} and TUMTraf\cite{zimmer2023tumtraf}, could also be explored.

{
    \small
    \bibliographystyle{ieeenat_fullname}
    \bibliography{main}
}

\clearpage
\setcounter{page}{1}
\maketitlesupplementary

\section*{A. Qualitative results}
\begin{figure}[ht]
    \centering
    \begin{subfigure}{0.45\textwidth}
        \centering
        \includegraphics[width=\linewidth]{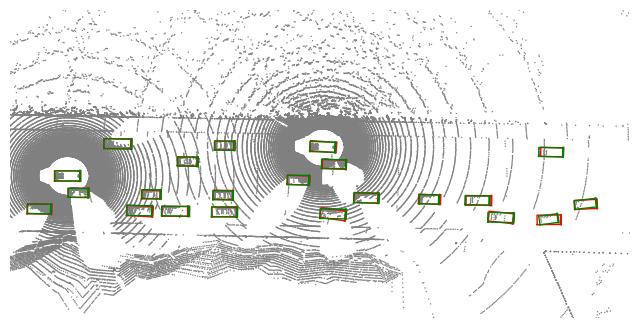}
        \caption{OPV2V}
        \label{fig:qua_opv2v}
    \end{subfigure}
    \vfill
    \begin{subfigure}{0.45\textwidth}
        \centering
        \includegraphics[width=\linewidth]{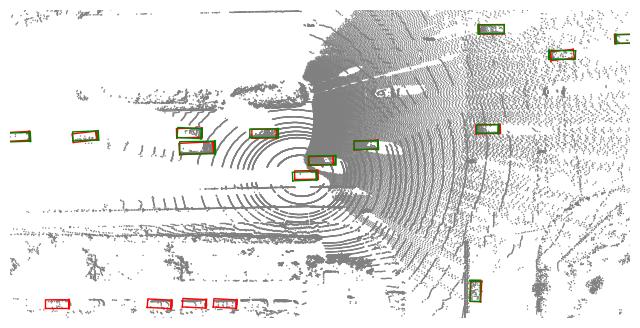}
        \caption{DairV2X}
        \label{fig:qua_dairv2x}
    \end{subfigure}
\vfill
    \begin{subfigure}{0.45\textwidth}
        \centering
        \includegraphics[width=\linewidth]{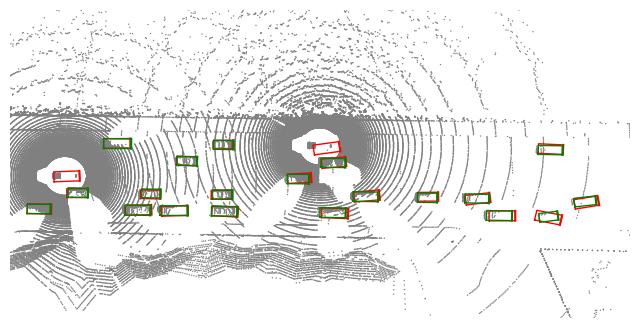}
        \caption{OPV2Vt}
        \label{fig:qua_opv2vt}
    \end{subfigure}
    \vfill
    \begin{subfigure}{0.45\textwidth}
        \centering
        \includegraphics[width=\linewidth]{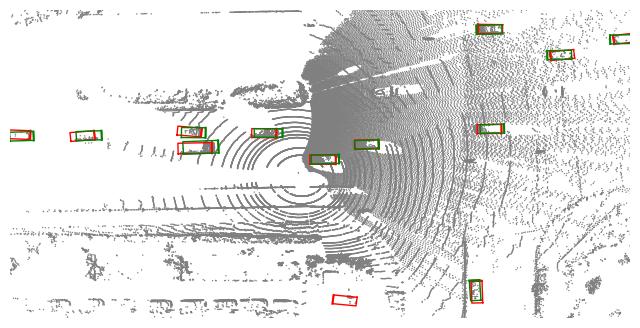}
        \caption{DairV2Xt}
        \label{fig:qua_dairv2xt}
    \end{subfigure}
    \caption{Qualitative results of COOD and TA-COOD. GT: green BBox, detection: red BBox.}
    \label{fig:qualitative}
\end{figure}

\begin{figure}[h]
    \centering
    \includegraphics[width=\linewidth]{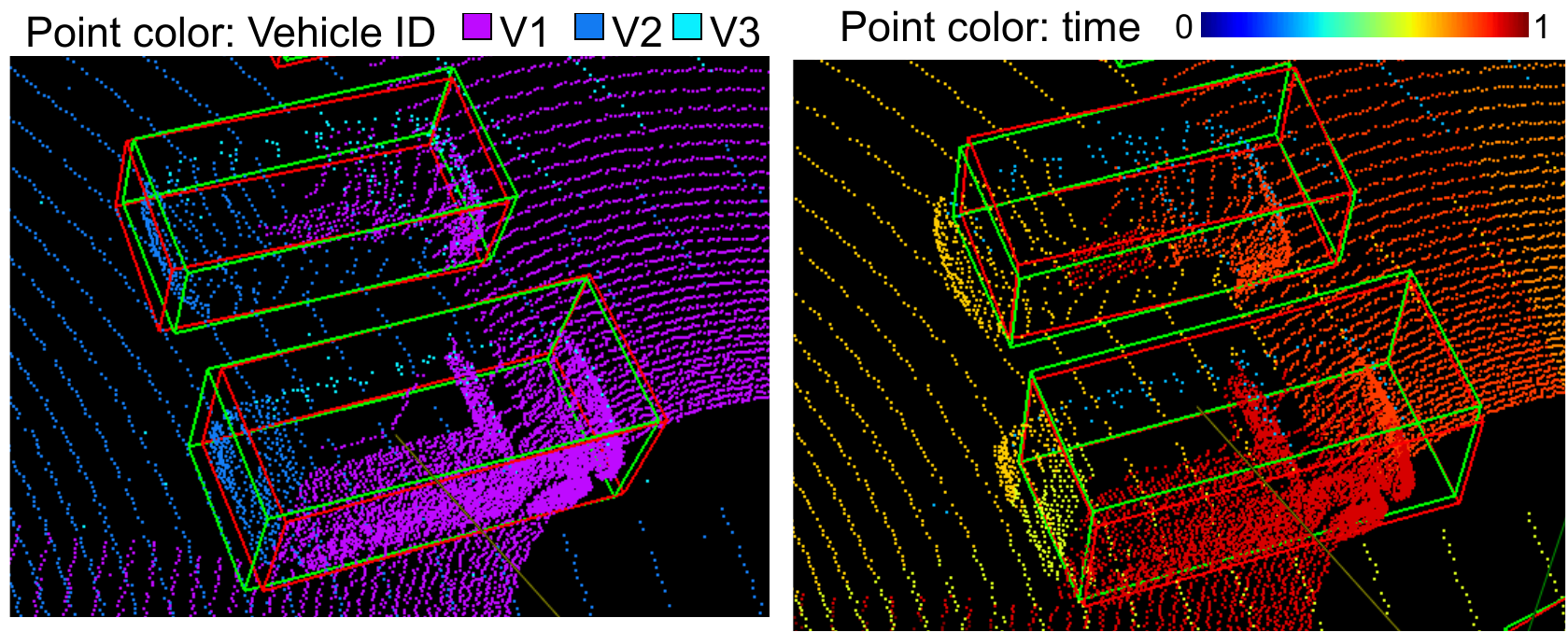}
    \vspace{-19pt}
    \caption{Qualitative result COOD (left) vs. TA-COOD (right).}
    \label{fig:qualitative_compare}
    \vspace{-15pt}
\end{figure}

\Cref{fig:qualitative} presents qualitative results for a sample frame from each dataset. The COOD results in \cref{fig:qua_opv2v,fig:qua_dairv2x} demonstrate that our framework, SparseAlign, successfully detects most vehicles with high overlap with the ground truth. In contrast, TA-COOD poses a greater challenge, leading to less accurate detections, as shown in \cref{fig:qua_opv2vt,fig:qua_dairv2xt}. However, the model performs better on OPV2Vt than on DairV2Xt. We attribute this to OPV2Vt being a simulated dataset with more precise ground truth, enabling more accurate learning of fine-grained temporal context.

\Cref{fig:qualitative_compare} illustrates the differences between COOD and TA-COOD. On the left, point clouds scanned by three IAs are visualized in different colors. The ground truth BBoxes (green) are perfectly aligned with the synchronized scanned points, allowing the model to focus solely on learning the geometric structure of the point clouds for accurate vehicle detection.
However, in TA-COOD, a more realistic setting, LiDAR points are typically scanned at different time points rather than being synchronized. In the right image, these asynchronized points are color-coded from blue (earlier scans) to red (later scans). Additionally, the ground truth bounding boxes are aligned to a future time frame rather than the exact geometry of the scanned points. This ground truth encourages the model to leverage temporal context to accurately predict vehicle positions in the near future, compensating for location errors caused by communication latency.

\section*{B. Free Space Augmentation (FSA)}
\begin{figure*}[t]
    \centering
    \includegraphics[width=0.8\linewidth]{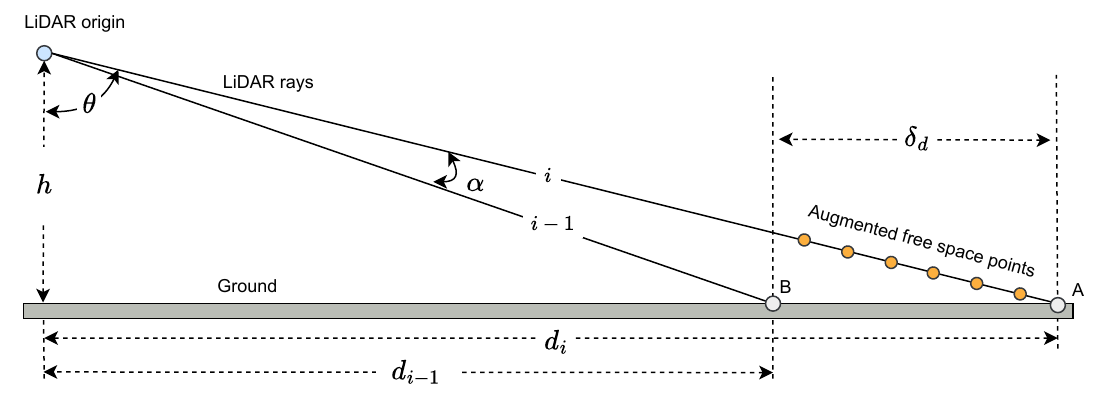}
    \caption{Free space augmentation.}
    \label{fig:fsa_geometry}
\end{figure*}

\begin{figure}[t]
\centering
    \includegraphics[width=0.45\textwidth]{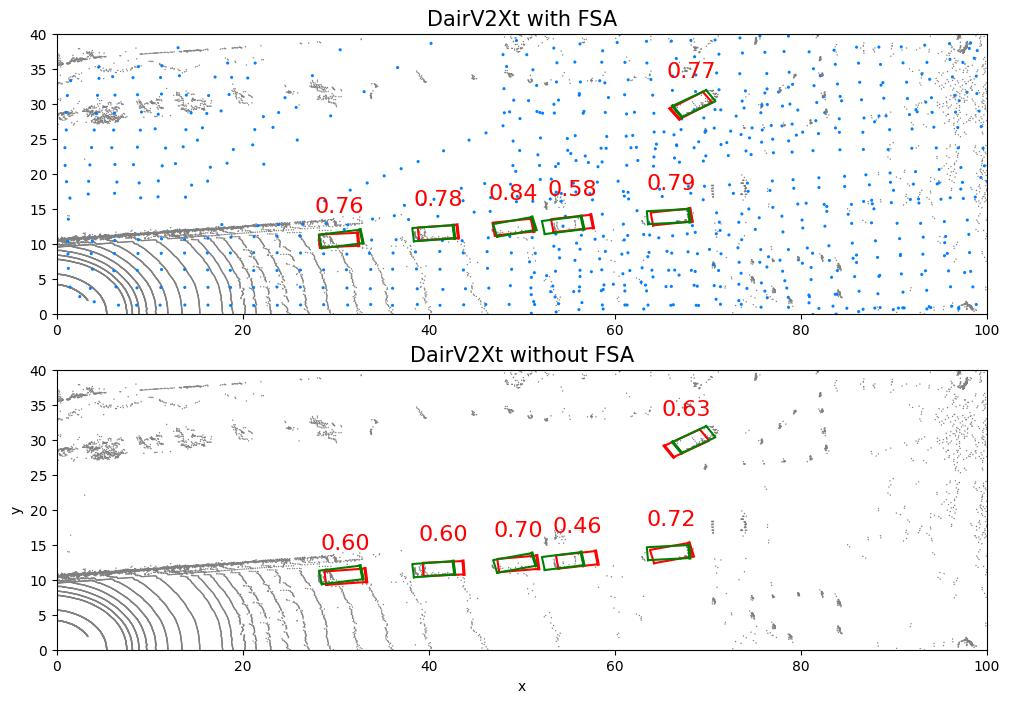}
    \caption{SparseAlign performance with and without FSA. The FSA points are in blue. Red texts are IoUs between the detected (red) and the ground-truth (green) BBoxes.}
    \label{fig:fsa}
\end{figure}

Free space augmentation (FSA) was first introduced in \cite{Yuan_gevbev2023} for BEV map-view semantic segmentation. Typically, point cloud data only capture reflections from obstacle surfaces. For example, in \cref{fig:fsa_geometry}, points A and B represent reflections from the ground surface. With only these measurements, we can confirm the presence of obstacles at these locations but have no information about the space between them. However, distinguishing between empty and unknown spaces is crucial in many applications. For instance, autonomous vehicles can safely navigate through known empty spaces but must avoid unknown areas where no measurements exist.

Each LiDAR ray provides information not only through its reflection point but also along its entire path, which consists of free space points indicating empty space. Augmenting sparse point clouds with additional points along LiDAR rays helps fill in these gaps. However, adding points along the entire ray path would be computationally prohibitive. Instead, as illustrated by the yellow points in \cref{fig:fsa_geometry}, it is sufficient to augment only the free space points between adjacent laser beams.

Mathematically, given the $i$-th LiDAR ray with ground distance $d_i$, LiDAR height relative to the ground $h$, and the inclination angle difference $\alpha$ between two adjacent rays (e.g., $\theta$ for the $i$-th ray), the gap distance between adjacent rays is calculated as:
\begin{equation} \delta_d = d_i - d_{i-1} = d_i - h\cdot\tan{(\arctan{\frac{d_i}{h}} - \alpha)} \end{equation}
Within this gap distance, we evenly distribute free space points to augment the original point cloud. These points not only convey information about empty spaces but also enhance connectivity between disjoint LiDAR scan rings. This, in turn, improves the convolutional layers' ability to learn spatial context over a larger receptive field.

\Cref{fig:fsa} compares SparseAlign’s performance with and without FSA. The results show that FSA slightly improves detection accuracy. Additionally, the augmented free space points (blue) incorporate timestamps computed based on their angles in the polar coordinate system. This enhances temporal context learning, particularly in distant regions where scan observations are sparse.

\section*{C. Dilation convolution}
In this paper, we demonstrated that \textit{SUNet} effectively mitigates ICF and CFM issues by expanding the receptive field through CEC layers. One might argue that dilated convolutional layers could serve a similar purpose. To investigate this, we conducted an additional experiment, modifying the dilation size of the first convolution layer in each \textit{SUNet} block from one to two. The results indicate that, in this case, dilated convolutions fail to effectively expand the receptive field or mitigate the ICF issue. Instead, they degrade local feature learning.
\begin{table}[h]
\centering
\begin{tabular}{c|cc|cc}
\toprule
Dilation size & \multicolumn{2}{c|}{OPV2V} & \multicolumn{2}{c}{DairV2X} \\ 
&\multicolumn{1}{c}{ AP0.5}    &  AP0.7&\multicolumn{1}{c}{ AP0.5}    &  AP0.7 \\ \hline 
2 &0.890  & 0.827  & 0.707 & 0.637\\ 
1 &0.924  & 0.885  & 0.773 & 0.638\\ 
\bottomrule
\end{tabular}
\caption{Comparison of normal sparse convolutions and dilation convolutions}
\label{tab:ablation_dil}
\end{table}

\section*{D. Gradients calculation schedule for efficient training}\label{sec:supp_grad}
\begin{figure}
    \centering
    \includegraphics[width=\linewidth]{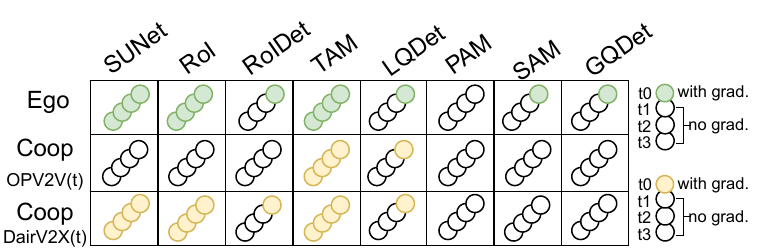}
    \caption{Schedules for gradient calculation.}
    \label{fig:grad_schedule}
\end{figure}
To optimize training efficiency, we schedule gradient calculations only for essential modules, as illustrated in \cref{fig:grad_schedule}. $t0$ is the newest frame and $t1$ to $t3$ is the historical frames. For the ego vehicle, we compute gradients for all modules in the latest data frame, except for \textit{PAM}, which is trained independently. Additionally, \textit{SUNet}, \textit{RoI}, and \textit{TAM} compute gradients across all frames to facilitate temporal feature learning.

In the OPV2V and OPV2Vt datasets, each IA is equipped with identical range-view sensors. Consequently, we disable gradient calculations for the first three modules in cooperative IAs, as their data exhibit the same feature patterns as those of the ego IA.

In contrast, in the DairV2X and DairV2Xt datasets, cooperative IAs are infrastructure-based, featuring different sensors and viewing angles from vehicles. To ensure effective learning from infrastructure data, gradient calculations remain enabled for \textit{SUNet}, \textit{RoI}, and \textit{TAM} in cooperative agents.

The final two modules pertain to the fusion process, which occurs exclusively in the ego vehicle. Therefore, no gradient calculations are performed on the cooperative side. Notably, we enable gradients for only one cooperative IA when necessary, irrespective of the total number of IAs.

\section*{E. Average precision on 3D metric}\label{sec:ap3d}
\begin{table}[th]
    \centering
    \begin{tabular}{c|c|c}
    \toprule
    & TransIFF\cite{TransIFF} & SparseAlign\\\hline
        OPV2V & - & 0.922/0.816  \\\hline
        DairV2X & 0.596/0.460 & 0.727/0.352\\\hline
        OPV2Vt & -& 0.898/0.703\\\hline
        DairV2Xt & -& 0.698/0.237\\\bottomrule
    \end{tabular}
    \caption{AP 3D at IoU thresholds of 0.5/0.7.}
    \label{tab:ap3d}
\end{table}
In the paper, we used BEV IoU thresholds to calculate the average precision. Here, we also report results based on 3D IoU thresholds for convenient comparison with other methods that use 3D AP as an evaluation metric, such as TransIFF\cite{TransIFF}. The results in \cref{tab:ap3d} indicate that our SparseAlign achieves higher AP at a 3D IoU threshold of 0.5 but lower AP at 0.7. We attribute this to the fact that the three alignment modules in SparseAlign consider only the $x$ and $y$ coordinates in geometric operations during the fusion process, without explicitly accounting for the $z$-axis.

\end{document}